\documentclass[letterpaper]{article} %
\usepackage{aaai2026}  %
\usepackage{times}  %
\usepackage{helvet}  %
\usepackage{courier}  %
\usepackage[hyphens]{url}  %
\usepackage{graphicx} %
\usepackage{natbib}  %
\usepackage{caption} %
\usepackage{algorithm}
\usepackage{algorithmic}
\usepackage{amssymb}
\usepackage{amsmath}
\usepackage{booktabs}
\usepackage{bbding}
\usepackage{pifont}
\usepackage{multirow}

\usepackage{bbding}
\usepackage{marvosym}

\usepackage{newfloat}
\usepackage{listings}
\DeclareCaptionStyle{ruled}{labelfont=normalfont,labelsep=colon,strut=off} %
\floatstyle{ruled}
\newfloat{listing}{tb}{lst}{}
\floatname{listing}{Listing}
\nocopyright

\title{E-VRAG: Enhancing Long Video Understanding with Resource-Efficient Retrieval Augmented Generation}
\author{
    Zeyu Xu\equalcontrib ,
    Junkang Zhang\equalcontrib,
    Qiang Wang\equalcontrib,
    Yi Liu\thanks{ Project Leader;
    \textsuperscript{\Letter}Corresponding Author.
  }\textsuperscript{\Letter}
}

\affiliations{
    Honor Device Co., Ltd~~~\\
    liuyi26@honor.com

}

\usepackage{bibentry}

\begin{document}

\maketitle

\begin{abstract}
Vision-Language Models (VLMs) have enabled substantial progress in video understanding by leveraging cross-modal reasoning capabilities. However, their effectiveness is limited by the restricted context window and the high computational cost required to process long videos with thousands of frames. Retrieval-augmented generation (RAG) addresses this challenge by selecting only the most relevant frames as input, thereby reducing the computational burden. Nevertheless, existing video RAG methods struggle to balance retrieval efficiency and accuracy, particularly when handling diverse and complex video content. Offline paradigms enable rapid retrieval by reusing pre-extracted static frame features, but they fail to capture fine-grained query-frame relationships, leading to suboptimal performance for tasks demanding diverse and detailed semantics. In contrast, online paradigms achieve higher retrieval accuracy via joint query-frame representations, but incur significant computational overhead as model size and video length increase. To address these limitations, we propose E-VRAG, a novel and efficient video RAG framework for video understanding. We first apply a frame pre-filtering method based on hierarchical query decomposition to eliminate irrelevant frames, reducing computational costs at the data level. We then employ a lightweight VLM for frame scoring, further reducing computational costs at the model level. Additionally, we propose a frame retrieval strategy that leverages the global statistical distribution of inter-frame scores to mitigate the potential performance degradation from using a lightweight VLM. Finally, we introduce a multi-view question answering scheme for the retrieved frames, enhancing the VLM's capability to extract and comprehend information from long video contexts. Experiments on four public benchmarks show that E-VRAG achieves about 70\% reduction in computational cost and higher accuracy compared to baseline methods, all without additional training. These results demonstrate the effectiveness of E-VRAG in improving both efficiency and accuracy for video RAG tasks.
\end{abstract}

\section{Introduction}

With the explosive growth of multimedia content, video understanding has become a key area in artificial intelligence. Vision-Language Models (VLMs) excel at cross-modal fusion, jointly modeling videos and semantics to enable richer context awareness and more complex reasoning. They have demonstrated superior performance in various video tasks, such as video retrieval, question answering (QA), and event detection~\cite{lin2024videollavalearningunitedvisual,wang2024internvideo2scalingfoundationmodels,li2024videochatchatcentricvideounderstanding}. However, VLMs struggle with long videos containing thousands of frames, mainly due to limited context windows and high computational demands. To address these challenges, some works focus on training VLMs with longer context windows~\cite{zhang2024longcontexttransferlanguage, chen2024longvilascalinglongcontextvisual, wang2024longllavascalingmultimodalllms}, which require large-scale video-text paired datasets and substantial computational resources. Other works aim to reduce visual tokens through token compression~\cite{song2024moviechatdensetokensparse, ren2024timechattimesensitivemultimodallarge}, which inevitably leads to the loss of fine-grained information and a decrease in performance on complex and detailed understanding.

Recently, Retrieval-Augmented Generation (RAG) has attracted significant attention in video understanding~\cite{jeong2025videoragretrievalaugmentedgenerationvideo}. The typical RAG framework retrievals the most relevant frames by computing relevance scores between numerous frames and the given query, and inputs these distilled key cues for deeper reasoning. By filtering data before inputting into VLMs, RAG significantly reduces computational overhead and context confusion, improving the adaptability of VLMs in video processing and understanding. 

\begin{figure}[t]
\centering
\includegraphics[width=1.02\columnwidth]{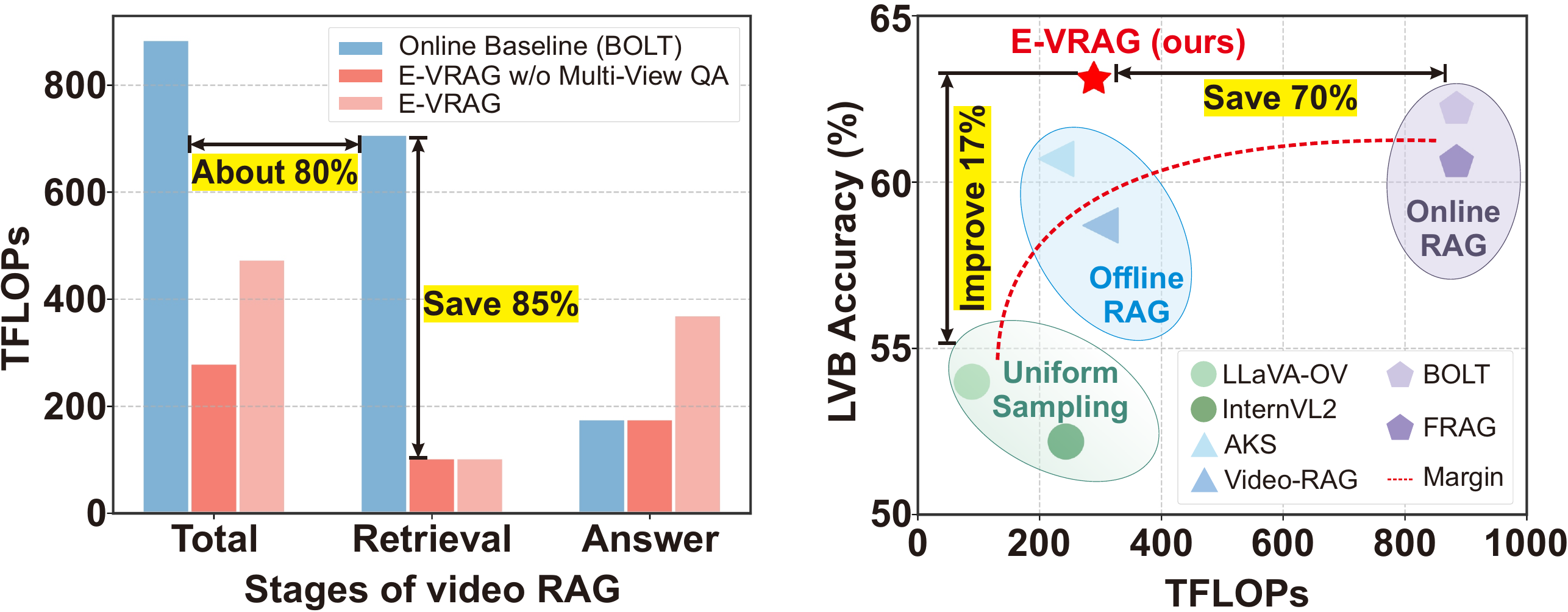} %
\caption{The computational analysis for stages of video RAG (left). The comparison of FLOPs between baselines and our E-VRAG (right).}
\label{fig:flops}
\end{figure}

However, video RAG faces an efficiency-accuracy trade-off. Offline video RAG methods prioritize efficiency by pre-extracting generic frame features, enabling reuse for each query~\cite{luo2024videoragvisuallyalignedretrievalaugmentedlong}. However, these static features often miss nuanced query-frame relationships, reducing retrieval quality for tasks with diverse semantics or granularity. In contrast, online video RAG methods prioritize accuracy by jointly modeling the relationship between each frame and the query, typically with a VLM~\cite{huang2025fragframeselectionaugmented, liu2025boltboostlargevisionlanguage}. However, their computational cost increases with both the model size and the number of frames, and to achieve comprehensive coverage, large models and a large number of frames are often employed, resulting in significant speed bottlenecks. As illustrated in the left of Figure~\ref{fig:flops}, when using one 7B model for video RAG, retrieval from 256 frames accounts for about 80\% of the total computation cost, greatly reducing efficiency and constraining practical applicability.

In this paper, we propose E-VRAG, which reduces retrieval computation at both data and model levels, and enhances accuracy through novel retrieval and QA methods, achieving significant margin improvements in both accuracy and efficiency, as shown in the right of Figure~\ref{fig:flops}. First, since only a small subset of frames are usually relevant, we introduce a pre-filtering method to coarsely filter frames, reducing unnecessary subsequent computation. Next, we utilize a lightweight VLM to efficiently compute the relationships between frames and queries, thereby further reducing computational overhead. To address potential performance drops from the lightweight VLM, we propose a retrieval method based on the global distribution of inter-frame scores, ensuring comprehensive retrieval. Finally, considering it is hard to extract all valuable cues from numerous retrieved frames by VLM at once, we design a multi-view QA scheme that iteratively performs QA and integrates feedback from different views, enhancing video understanding with the given query. Experiments show that E-VRAG achieves lower computational cost and higher accuracy in a fully training-free setting.
Our main contributions can be summarized as:

\begin{itemize}
    \item We propose a frame pre-filtering method based on hierarchical query decomposition and score frames with a lightweight VLM to reduce the computational overhead of video RAG at both data and model levels.
    \item We proposed a frame retrieval method based on the global distribution of inter-frame scores and a multi-view QA scheme to enhance the accuracy of retrieval and response.
    \item Our proposed E-VRAG reduces computation by up to 70\% while maintaining or surpassing the accuracy of baseline models on four public benchmarks, under the training-free setting.
\end{itemize}

\section{Related Works}

\subsection{Large Video-Language Models}
Open-source VLMs, such as LLaVA-OneVision~\cite{li2024llavaonevisioneasyvisualtask}, NVILA~\cite{liu2025nvilaefficientfrontiervisual}, InternVL2~\cite{chen2025expandingperformanceboundariesopensource}, and Qwen2.5VL~\cite{bai2025qwen25vltechnicalreport}, have greatly advanced visual understanding and show powerful capabilities across diverse general visual tasks. According to Scaling Laws~\cite{kaplan2020scalinglawsneurallanguage}, to improve the visual understanding capabilities, the parameters of VLMs are gradually expanded to 72B or even larger, resulting in a higher cost for application especially for videos. Thus, more efficient VLMs specifically designed for videos have emerged. Some works focus on the alignment of video representations with other modalities, such as VideoLLaVA~\cite{lin2024videollavalearningunitedvisual}, InternVideo2~\cite{wang2024internvideo2scalingfoundationmodels}, and VideoChat~\cite{li2024videochatchatcentricvideounderstanding}. Some works concentrate on the extraction and compression of spatiotemporal information from videos, such as Video-ChatGPT~\cite{maaz2024videochatgptdetailedvideounderstanding} and VideoLLaMA2~\cite{cheng2024videollama2advancingspatialtemporal}.
There are also some works focus on data and training, enhancing video understanding capabilities through synthetic data and better training schemes, such as ShareGPT4Video~\cite{chen2024sharegpt4videoimprovingvideounderstanding}, LLaVA-Video~\cite{zhang2024videoinstructiontuningsynthetic}, and VideoLLaMA3~\cite{zhang2025videollama3frontiermultimodal}. Although the video understanding capabilities of existing models have significantly improved, there remain substantial challenges in understanding long videos.

\begin{figure*}[t]
\centering
\includegraphics[width=\textwidth]{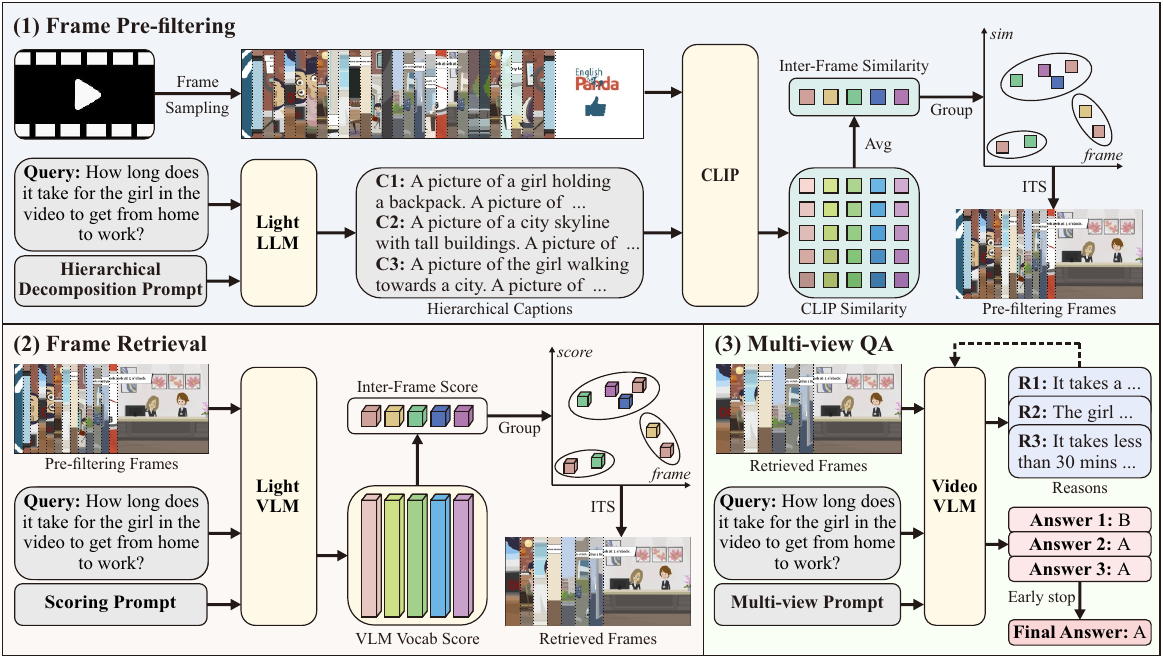} %
\caption{
The overall workflow of our E-VRAG.
 First, the frame pre-filtering stage quickly filters relevant frames using query decomposition and inter-frame similarity. Next, the frame retrieval stage accurately retrievals frames using a lightweight VLM and inter-frame scores. Finally, the multi-view QA stage enhances the answer from multiple views.}
\label{fig1}
\end{figure*}

\subsection{Long Video Understanding}
Current VLMs often struggle with long video understanding due to limited context length and low efficiency. Recent works have provided various solutions to address these challenges. Some works try to extend the context length of VLMs through training, enabling the processing of more frames, such as LongVA~\cite{zhang2024longcontexttransferlanguage}, LongVILA~\cite{chen2024longvilascalinglongcontextvisual}, and LongLLaVA~\cite{wang2024longllavascalingmultimodalllms}. However, training a long-context VLM is costly in terms of data and resources. Some works focus on compressing and refining visual information to adapt to VLMs, such as MovieChat~\cite{song2024moviechatdensetokensparse}, TimeChat~\cite{ren2024timechattimesensitivemultimodallarge}, VideoChat-Flash~\cite{li2025videochatflashhierarchicalcompressionlongcontext}, LongVU~\cite{shen2024longvuspatiotemporaladaptivecompression}, Video-XL2~\cite{qin2025videoxl2longvideounderstandingtaskaware}. While compression improves efficiency, it can also lead to information loss, potentially limiting the performance of VLM on fine-grained or complex tasks. 
\subsection{Video Understanding with RAG}
Combining RAG with VLMs for long video understanding has recently gained significant attention. 
Video RAG can be typically categorized as offline or online methods. 
Offline methods extract frame features independently of queries, enabling the reuse of frame features for efficient scoring and retrieval across different queries, such as VideoRAG~\cite{jeong2025videoragretrievalaugmentedgenerationvideo}, 
Video-RAG~\cite{luo2024videoragvisuallyalignedretrievalaugmentedlong}, 
Q-Frame~\cite{zhang2025qframequeryawareframeselection}, 
and MemVid~\cite{yuan2025memoryenhancedretrievalaugmentationlong}. 
However, the pre-extracted static frame features are hard to handle diverse and fine-grained queries, resulting in decreased retrieval accuracy. 
Online methods use models to jointly extract relationship between each frame and query, enabling more detailed and accurate retrieval, such as FRAG~\cite{huang2025fragframeselectionaugmented}, BOLT~\cite{liu2025boltboostlargevisionlanguage},  RAG-Adapter~\cite{tan2025ragadapterplugandplayragenhancedframework}, GenS~\cite{yao2025generativeframesamplerlong}, ViaRL~\cite{xu2025viarladaptivetemporalgrounding}, and Frame-voyager~\cite{yu2025framevoyagerlearningqueryframes}. However, online methods require recomputing the matching database for frames with each input query, which significantly reduce efficiency. Additionally, some methods achieve enhancement of video RAG through iterative querying, such as VideoAgent ~\cite{wang2024videoagentlongformvideounderstanding} and MoReVQA~\cite{min2025morevqaexploringmodularreasoning}. 
Overall, the existing online and offline video RAG methods face the challenge of balancing efficiency and accuracy. To reach the Pareto frontier, we focus on reducing the computation costs while maintaining performance  across the entire video RAG process, aiming for optimal efficiency and accuracy.

\section{Methods}
In this section, we introduce E-VRAG, highlighting efficiency enhancements at both data and model levels, retrieval accuracy improvement via inter-frame grouping, and answer refinement through multi-view integration.
\subsection{Overview}
The general video understanding can be formulated as: 
\begin{equation}
\mathcal{A} = \mathrm{VLM}\left(\mathcal{F},\;  \mathcal{Q}\right),
\end{equation}
where the input $\mathcal{F} =  \left\{ \mathrm{Frame}_i \right\}_{i=1}^N$ is $N$ frames in video and generally uniform sampled, $\mathcal{Q}$ is the qurey, and $\mathcal{A}$ is the answer. Unrestrictedly increasing $N$ may exceed the context length limitation of VLM and obscure key information. The RAG-based video understanding methods address this by retrieving $\mathcal{R}$ and inputting only the most relevant frames:
\begin{equation}
\mathcal{A} = \mathrm{VLM}\left( \mathcal{R}\left( \mathcal{F}, \mathcal{Q} \right),\; \mathcal{Q} \right).
\end{equation}

Clearly, the key to video RAG is efficiently and accurately locating relevant frames and comprehensively understanding their content. Thus, we propose E-VRAG, an efficient video RAG that enhances video understanding. As shown in Figure~\ref{fig1}, E-VRAG has three stages: frame pre-filtering hierarchically matches queries to frames and filters out irrelevant frames; frame retrieval uses a lightweight VLM and inter-frame scores for efficiently and accurate retrieval; multi-view QA answers queries from multiple views to boost understanding.

\subsection{Frame Pre-filtering}
\subsubsection{Similarity with Hierarchical Query Decomposition.}
We think most frames are only weakly related to the query and can be quickly filtered by coarse multi-modal alignment, without detailed analysis. However, pre-trained CLIP-like models align images with captions, which often has a semantic gap with queries, since relevant information may be implicit in queries and not explicitly stated as in captions, leading to retrieval failures.

Therefore, we decompose the query into three levels and transform them into captions that are more suitable for image matching: 
\begin{equation}
\mathcal{C}_{H} = \mathrm{LLM}\left(\mathcal{Q},\; \mathcal{P}_{H}\right) = \left\{ \mathcal{C}_{entity},\, \mathcal{C}_{know},\, \mathcal{C}_{causal} \right\},
\label{eq:decomp}
\end{equation}
where $\mathcal{P}_{H}$ is the hierarchical decomposition prompt. $\mathcal{C}_{entity}$ focuses on directly describable entities mentioned in the query, which are the simplest cases and can be directly translated into image captions (e.g., cats, dogs). $\mathcal{C}_{know}$ focuses on entities or abstract concepts that cannot be directly described, which need the LLM to serve as a knowledge base, leveraging external knowledge to convert these entities into visual captions (e.g., describing New York as a city with skyscrapers or referencing the Statue of Liberty). $\mathcal{C}_{causal}$ focuses on causal and logical relationships within the query, also requiring the LLM to serve as a knowledge base to supplement the relevant captions of events that are not explicitly mentioned. 

We utilize CLIP~\cite{radford2021learningtransferablevisualmodels} to compute the similarity $\mathcal{S}$ between frames and decomposed captions:
\begin{gather}
F_\mathrm{c} = \mathrm{CLIP}_\mathrm{t}\left(\mathcal{C}_{H}\right) ,\quad F_\mathrm{v} = \mathrm{CLIP}_\mathrm{v}\left(\mathcal{F}\right),  \\
\mathcal{S} = \frac{1}{C} \sum_{i=1}^{C} \frac{ F_\mathrm{c}^{i} \cdot F_{\mathrm{v}} }{ \| F_\mathrm{c}^{i} \| \; \| F_{\mathrm{v}} \| }\in \mathbb{R}^{N},
\end{gather}
where $\mathrm{CLIP}_\mathrm{t}$ and $\mathrm{CLIP}_\mathrm{v}$ are the text and vision encoders of CLIP, respectively, $C$ is the total caption number of $\mathcal{C}_{H}$, and $\cdot$ is the inner product. This stage is computationally efficient, as the query decomposition can be combined into a single LLM inference, and the image features within the same video can be reused. 

\subsubsection{Filtering with Inter-frame Similarity.}
Typically, the Top-K frames with the highest similarity scores are selected, which works well only when features are highly distinguishable. Otherwise, the selected frames are temporally adjacent, oversampling major events and undersampling others. 

Therefore, we consider the global distribution of inter-frame similarity, and group frames to ensure higher similarity within groups than between groups. We group frames through clustering with a specific temporal constraint: only frames that are both temporally adjacent and similar are grouped. The algorithmic details are provided in the supplementary materials. By incorporating temporal relationships, it can be beneficial for causal understanding, as similar events at different times are analyzed separately rather than merged.

Then, we sample the frames within each group individually to control redundancy while preserving representativeness. The sampling method is Inverse Transform Sampling (ITS)~\cite{liu2025boltboostlargevisionlanguage}. For each group $g$, it uniformly samples frames $\mathcal{F}_g$ based on the inverse function of the cumulative distribution of similarity $\mathcal{S}_g'$ to avoid redundant dense sampling:
\begin{equation}
\begin{gathered}
\mathcal{F}_g' = \left\{ \mathcal{F}_g(i) \;\middle|\; i={\left \lfloor \mathcal{S}_g'(\frac{j}{M_g})\right \rfloor }  \right\},\\
 \quad g = 1,\ldots,G;\; j = 1,\ldots,M_g;\;M_g = \frac{N_g}{N}M,
\end{gathered}
\end{equation}
 where $N_g$ is the frame number of $\mathcal{F}_g$, $M$ is the number of frames expected to be selected, $G$ is the number of groups, and $\lfloor \rfloor$ is the floor operation.

\subsection{Frame Retrieval}
\subsubsection{Scoring with Lightweight VLM.}
For the potentially relevant frames obtained after pre-filtering, we further utilize the multi-modal understanding capabilities of the VLM to evaluate their relevance to the query. Specifically, each frame is paired with the query and input to the VLM using a binary relevance judgment (${yes}$/${no}$) prompt. The VLM generates a binary response based on the instruction, and we use the full vocabulary probability distribution $P_\text{all}\in \mathbb{R}^{v}$ of the answer word as the relevance score for retrieval, where $v$ is the vocabulary size of the VLM. Since this stage requires multiple VLM inferences, we use only a lightweight VLM to ensure efficiency.

\begin{table*}[t]
\centering
\small
\renewcommand{\arraystretch}{1.15}
\begin{tabular}{lcccccccc}
\toprule
\textbf{Method} & \textbf{Retrieval} & \textbf{Answer} & \textbf{TFLOPs} & \textbf{\#Frames} & \textbf{Video-MME} & \textbf{MLVU} & \textbf{LVB} & \textbf{NextQA} \\
\midrule
\multicolumn{9}{l}{\textbf{Fundamental Methods}} \\
\midrule
VideoChat2~\cite{li2024mvbench}   
& -- & 7B & -- & -- & 39.5 & 44.5 & 39.3 & -- \\
VideoLLaMA2~\cite{cheng2024videollama2advancingspatialtemporal}    
& -- & 7B & -- & -- & 47.9 & --   & --   & -- \\
InternVL2$^{*}$~\cite{chen2024far}       
& -- & 8B & 243 & 64 & 56.6 & 60.7 & 52.2 & 80.6 \\
LLaVA-OV$^{*}$~\cite{li2024llavaonevisioneasyvisualtask}        
& -- & 7B & 89 & 32 & 57.4 & 61.8 & 54.0 & 78.8 \\
Qwen2.5VL$^{*}$~\cite{bai2025qwen25vltechnicalreport}       
& -- & 7B & 130 & 32 & 62.1 & 59.6 & 58.1 & 81.6 \\
LLaVA-Video$^{*}$~\cite{zhang2024videoinstructiontuningsynthetic}     
& -- & 7B & 177 & 64 & 64.3 & 69.5 & 61.2 & \underline{83.8} \\
\midrule
\multicolumn{9}{l}{\textbf{Offline Video RAG Methods}} \\
\midrule
Video-RAG~\cite{luo2024videoragvisuallyalignedretrievalaugmentedlong}         
& 0.3B & 7B & -- & 64  & -- & \textbf{72.4} & 58.7  & -- \\
VideoAgent~\cite{wang2024videoagentlongformvideounderstanding}          
& 8B & GPT-4 & -- & --  & -- & -- & -- & 71.3 \\
Goldfish~\cite{ataallah2024goldfishvisionlanguageunderstandingarbitrarily}             
& 7B & 7B & -- & --  & 28.9 & 37.3 & -- & -- \\
MemVid~\cite{yuan2025memoryenhancedretrievalaugmentationlong}               
& 7B & 7B & -- & 128  &  63.7 & 58.1 & -- & -- \\
AKS$^{*\dagger}$~\cite{tang2025adaptivekeyframesamplinglong}        
& 0.3B & 7B & 50+177 & 64 & 64.3 & 69.3 & 60.7 & 83.3 \\
\midrule
\multicolumn{9}{l}{\textbf{Online Video RAG Methods}} \\
\midrule
Frame-voyager~\cite{yu2025framevoyagerlearningqueryframes}    
& 7B & 7B & -- & 8  & 57.5 & 65.6 & --   & 73.9 \\

FRAG$^{*\dagger}$~\cite{huang2025fragframeselectionaugmented}       
& 7B & 7B & 708+177 & 64 & 63.7 & 69.2 & 60.6 & 82.5 \\
BOLT$^{*\dagger}$~\cite{liu2025boltboostlargevisionlanguage}       
& 7B & 7B & 708+177 & 64 & \underline{64.6} &  \underline{70.3} & \underline{62.2} & 83.2 \\

\textbf{E-VRAG (ours)} & 2B & 7B & 103+372 & 64 & \textbf{65.4} & 70.2 & \textbf{63.1} & \textbf{84.0} \\
\bottomrule
\end{tabular}

\caption{Comparison of E-VRAG with SOTA video understanding methods. *: local implementation using official code/weights for fair comparison,  $\dagger$: 256 candidate frames uniformly sampled per video for fair comparison, values before/after ``+": TFLOPs for retrieval/answer, bold: highest, underline: second highest. For more details, see the supplementary materials.}
\label{tab:main_results}
\end{table*}

\subsubsection{Retrieval with Inter-frame Probability.}
Given the capability limitations of lightweight VLM in scoring, we continue to use the grouping and sampling strategy from the frame pre-filtering stage to maintain retrieval quality. Specifically, frames are grouped based on their vocabulary probability $P_\text{all}$, under the assumption that frames with more similar $P_\text{all}$ distributions are likely to be more correlated and redundant. For each group, we continue to employ ITS for frame retrieval, modifying the similarity to the scores generated by $P_\text{all}$ while keeping others unchanged. We consider two scoring strategies: the first uses only the probability of word ${yes}$, i.e., $P_\text{all}\left(yes\right)$, while the second incorporates both words ${yes}$ and ${no}$, i.e., $\frac{P_\text{all}\left({yes}\right)}{P_\text{all}\left({yes}\right) + P_\text{all}\left({no}\right)}$. The retrieved frames $\mathcal{F}_r$ are subsequently utilized to answer the query.

\subsection{Multi-view QA}
A single VLM inference may struggle to fully extract all information from multiple frames in $\mathcal{F}_r$, especially fine-grained details. Enhancing by supplementing information through multi-round reasoning has demonstrated effectiveness. However, sequential aggregation reasoning within a single chain, where answer is generated solely based on the final inference, is prone to error accumulation in intermediate steps and may lack robustness. 

Thus, we propose multi-view QA, in which each round attempts to reason and answer  query from a distinct view, and answers are aggregated and complemented in parallel to produce a more comprehensive response. Specifically, in the $t$-th round of QA, the VLM generates a reason  $\mathcal{R}_t$  and an answer $\mathcal{A}_t$ based on the input frame  $\mathcal{F}_r$, the query $\mathcal{Q}$, as well as the reasons $\mathcal{R}_{< t}$ and answers $\mathcal{A}_{< t}$  from previous rounds, which serve as additional supplementary contexts: 
\begin{equation}
(\mathcal{R}_t,\mathcal{A}_t) = \mathrm{VLM}
(\mathcal{F}_r,\mathcal{Q},\mathcal{R}_{< t},\mathcal{A}_{< t}),\; t = 2, \dots, T, \\
\end{equation}
where $T$ is the total number of views. Notably, in the first view, only query and frames are input. In this process, the VLM is required to explicitly state the view used to analyze the frames and answer the query in its generated reasons. Additionally, it is prompted to respond from a view that differs from those employed in previous rounds. By incorporating feedback from multiple views, the VLM progressively refines its understanding of the query and frames.

To enhance efficiency, we introduce an early stopping strategy, whereby the process is terminated if answers generated in two consecutive rounds are identical. A voting mechanism is utilized to determine the final answer from all answers generated during the multi-view QA process. In cases where multiple answers have the same count, the answer generated latest among them is chosen as the final answer:
\begin{equation}
\begin{gathered}
\mathcal{A}^* = \mathrm{Vote}(\mathcal{A}_1,\dots,\mathcal{A}_k),\\ 
k = \min\left(T,\ \min\{t \mid \mathcal{A}_t = \mathcal{A}_{t-1},\ t \geq 2\}\right).
\end{gathered}
\end{equation}

\section{Experiments}
In this section, we present key experiments to validate the effectiveness of E-VRAG against various SOTA video understanding methods, particularly video RAG methods, on four benchmarks. Additional experiments are provided in the supplementary materials.
\subsection{Experimental Setup}
\subsubsection{Benchmarks}
We evaluated our method on four widely-used video benchmarks: Video-MME~\cite{fu2025video}, LongVideoBench~\cite{wu2024longvideobench}, MLVU~\cite{zhou2024mlvu}, and NextQA~\cite{xiao2021next}. Video-MME comprises a diverse collection of short, medium, and long videos, comprising a total of 900 videos and 2,700 questions, with an average video duration exceeding 8 minutes. Both LongVideoBench (LVB) and MLVU are specifically designed to assess models’ capabilities in understanding long videos, with an average video duration greater than 8 minutes. NextQA consists of shorter video clips, with an average duration of only 0.8 minutes. 
\subsubsection{Baseline Models}
We compare our E-VRAG with three types of baselines for evaluation. The first type includes fundamental methods that are general for visual understanding or video understanding without RAG. The second type includes offline video RAG methods that extract information of frames and queries separately by retrieval models for RAG. The third type includes online video RAG methods that extract information jointly by retrieval models for RAG. Through the above comparisons, the efficiency and accuracy of our EVIRAG can be comprehensively demonstrated.
\subsubsection{Implementation Details}
We conducted all experiments on 8 NVIDIA A800 80G GPUs without training. Following the experimental protocol in~\cite{liu2025boltboostlargevisionlanguage,huang2025fragframeselectionaugmented}, we uniformly sample 256 candidate frames and retrieve 64 frames with dynamic resolution of 1 to generate answers. For accuracy reporting, we use consistent settings across four datasets: a group number of 26, a scoring type of two words, a multi-view number of 2. The impact and select reason of these main hyperparameters are discussed in the ablation study and supplementary materials. The pre-filtering stage acts as a middle step between uniform sampling and retrieval. Thus, the number of filtered frames is set to 128, accounting for 50\% of candidate frames and twice the retrieval stage frames. Its group number is also set to twice that of the retrieval stage to maintain proportionality. The lightweight LLM used for query decomposition is Qwen3-1.7B~\cite{qwen3technicalreport}.

\subsection{Main Results}
\subsubsection{Accuracy Comparison with SOTA Methods}
Table~\ref{tab:main_results} presents a comparison between our method and three types of baseline methods. It can be find that the performances of most general fundamental methods are not as good as video RAG methods, especially in long video benchmarks like LVB or MLVU. Among these fundamental methods, the LLaVA-Video has the highest accuracy across four benchmarks, thus we use it as the answer model and perform comparisons with other SOTA video RAG methods equally. The accuracy of the offline methods is rarely higher than that of the online methods across the four benchmarks, which highlights the advantage of the online methods in accuracy. Our E-VRAG achieves the highest accuracy on three benchmarks comparing with other baselines, with an accuracy improvement of 0.8\% in Video-MME, 0.9\% in LVB, and 0.2\% in NextQA, comparing with the highest accuracy of all baselines. 

To comprehensively validate the generalization of E-VRAG, we further use LLaVA-OV and Qwen2.5VL as answer models and apply E-VRAG to them. The results are shown in the supplementary materials, where accuracy increases after apply E-VRAG. These results demonstrate that our E-VRAG is applicable to multiple fundamental models and provides consistent improvements.

\begin{table}[t]
\centering
\begin{tabular}{ccccccccc}
\toprule
FP.&FR.&QA.& TFs &\textbf{VE} & \textbf{MU} & \textbf{LB} & \textbf{NA} \\
\midrule
-- & -- & -- &  177 &64.3 & 69.5 & 61.2& 83.8 \\
\checkmark & -- & -- & 178 &64.5& 69.2&  60.4& 83.5\\
-- & \checkmark & -- &  381 &65.0&  \textbf{70.4}&  63.0& 83.7\\
\checkmark & \checkmark & -- & 280 &65.0 & 70.0&  \textbf{63.4}&  83.6  \\
\checkmark & \checkmark & \checkmark &  475 &\textbf{65.4}&  70.2&  63.1&  \textbf{84.0}\\
\bottomrule
\end{tabular}
\caption{The ablation results of three stages in  Efficient-VRAG. FP.: frame pre-filting, FR.: frame retrieval, QA.: multi-view QA, TFs: TFLOPs, VE: Video-MME, MU: MLVU, LB: LVB, NA: NextQA. The same abbreviation is used in the following tables to represent the same meaning.}
\label{tab:global_ablation}
\end{table}

\begin{table}[t]
\centering
\begin{tabular}{ccccc}
\toprule
Query Decom. & \textbf{VE} & \textbf{MU} & \textbf{LB}& \textbf{NA} \\
\midrule
 -- & 64.9& \textbf{70.3}& 61.6& 83.5\\
\checkmark&  \textbf{65.0} & 70.0&  \textbf{63.4}&  \textbf{83.6}\\
\bottomrule
\end{tabular}

\caption{The ablation results of query decomposition. Query Decom.: query decomposition.}
\label{tab:Pre-filtering}
\end{table}

\begin{figure*}[t]
\centering
\includegraphics[width=\textwidth]{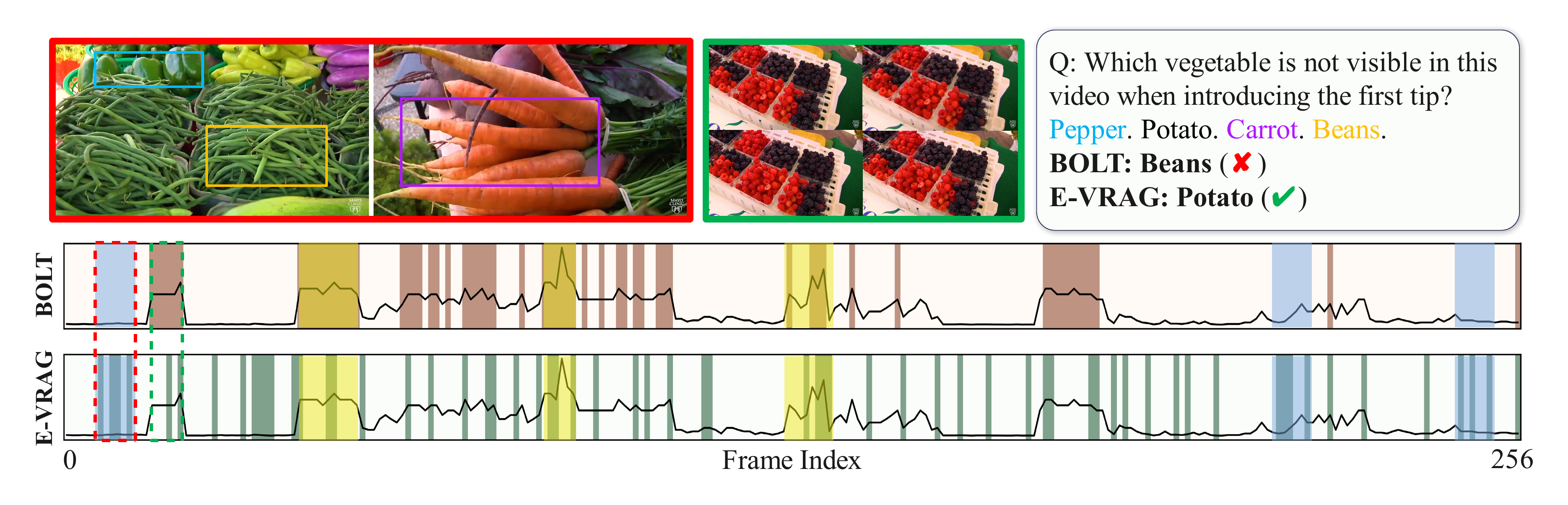} %
\caption{The visualization comparison with E-VRAG and online video RAG baseline.}
\label{fig:score}
\end{figure*}

\begin{table}[t]
\centering
\begin{tabular}{cccccc}
\toprule
FPGR.& FRGR.& \textbf{VE} & \textbf{MU} & \textbf{LB} &  \textbf{NA}  \\
\midrule
 -- &  -- & 62.5 & 69.4 & 60.1& 82.6\\
 \checkmark &  -- & 64.3 & \textbf{70.3}&  60.5&  83.0  \\
\checkmark&  \checkmark& \textbf{65.0} & 70.0&  \textbf{63.4}&  \textbf{83.6}  \\
\bottomrule
\end{tabular}

\caption{The ablation results of retrieval with inter-frame grouping. FPGR.: group retrieval in frame pre-filtering, 
FRGR.: group retrieval in frame retrieval.}
\label{tab:Inter_frame}
\end{table}

\begin{table}[t]
\centering
\begin{tabular}{cccccc}
\toprule
Views & TFLOPs & \textbf{VE} & \textbf{MU} & \textbf{LB}  & \textbf{NA}\\
\midrule
1 & 280&  65.0& 70.0&  \textbf{63.4}& 83.6\\
2 & 475&  65.4& 70.2&  63.1& 84.0\\
3 & 688&  65.4& \textbf{70.3}&  63.0& \textbf{84.1}\\
4 & 919& \textbf{65.5}& \textbf{70.3}&  63.0& \textbf{84.1}\\
5 & 1168& \textbf{65.5}& \textbf{70.3}&  63.2& \textbf{84.1}\\
\bottomrule
\end{tabular}

\caption{The accuracy of different view number in multi-view QA.}
\label{tab:Multi-viewQA}
\end{table}

\subsubsection{Efficiency Comparison with SOTA Methods}
Table~\ref{tab:main_results} shows the computation costs each method used. All answer models have a similar parameter size of 7B or 8B, so efficiency differences mainly due to the retrieval models. Fundamental methods uniformly sample frames without retrieval, resulting in minimal computational cost but the lowest accuracy. Offline and online RAG methods uses model to compute retrieval clues, which improves accuracy but increases computational load. Moreover, although the accuracy of online methods is higher than offline methods, the cost of increased computation outweighs its benefits. Our E-VRAG method incorporates online retrieval similar to FRAG and BOLT, but pre-filters irrelevant frames and uses a lightweight VLM, enhancing efficiency at both the data and model levels. Thus, it achieving about 50\% reduction in computational cost with even higher accuracy compared with baseline online methods. Moreover, according to the results in Table~\ref{tab:Multi-viewQA}, our E-VRAG can be more efficiency by replace the multi-view QA with a single-view QA, achieving about 70\% reduction in computational cost and still maintains good performances. These results demonstrate that combining a compact retrieval model with an appropriate frame retrieval method is an effective way to achieve a balance between efficiency and accuracy.

\subsection{Quantitative Comparison}
\subsubsection{Global Ablation}
We conducted ablation studies on three stages of E-VRAG, as shown in Table~\ref{tab:global_ablation}. Omitting all stages equals uniform sampling, which is computationally efficient but has low accuracy. Using only frame pre-filtering corresponding to the offline RAG, where decomposed queries are matched with static frame features and frames are retrieved based on coarse-grained understanding. It offers limited accuracy gains, likely due to its tendency to retrieve incorrect or redundant frames. Thus, it is more suitable for coarse filtering rather than direct retrieval. Using only frame retrieval corresponding to the online RAG, which models the relationship between queries and frames in detail and achieves better accuracy than previous methods. However, despite employing a lightweight VLM, the computational cost remains high, as inference is required for each frame-query pair. Combining above two stages leverages the strengths of each, resulting in an approximately 30\% reduction in computational cost while maintaining comparable performance, thus achieving a balance between accuracy and efficiency. Incorporating multi-view QA typically improves accuracy, but at the cost of higher computation. A more detailed analysis of three stages is presented in the following section.

\subsubsection{Analysis of Query Decomposition}
We evaluated the effectiveness of query decomposition using the ablation study shown in Table~\ref{tab:Pre-filtering}, where only the query decomposition in frame pre-filtering is replaced by directly inputting the original query into CLIP to compute text features for retrieval. After removing it, the accuracy on Video-MME, LVB and NextQA decreased, while that on MLVU increased. Among these benchmarks, the improvement brought by query decomposition is most significant on LVB, which may because queries in LVB are relatively long and contain more information, making them less suitable for retrieval using a single short caption as recommended by CLIP. 

\subsubsection{Analysis of Inter-frame Group Retrieval}
We evaluate the effectiveness of inter-frame group retrieval through the ablation study shown in Table~\ref{tab:Inter_frame}, where the inter-frame group retrieval in frame pre-filtering (FPGR.) and frame retrieval (FRGR.) stages is progressively removed, and replaced with Top-K retrieval. Removing it in both stages has the lowest accuracy, even lower than the basic LLaVA-Video with uniform sampling shown in Table~\ref{tab:main_results}, which indicates that Top-K selection is likely to be redundant and incomplete. After incorporating inter-frame group retrieval into both the frame pre-filtering and retrieval stages, the accuracy on almost all benchmarks increased compared to the previous results, demonstrating its effectiveness at each stage. The decrease on MLVU at frame retrieval stage may partially attributed to the use of a unified group number hyperparameter. However, the overall performance still surpasses that of the methods without group retrieval. 
We also comprehensively analyze how group number and score type affect inter-frame group retrieval in the supplementary materials.
\subsubsection{Analysis of Multi-view QA}
The accuracy results for different numbers of views are shown in Table~\ref{tab:Multi-viewQA}. For Video-MME, MLVU, and NextQA, accuracy consistently improves as the number of views increases, demonstrating the effectiveness of the multi-view QA. However, as the number of views continues to grow, accuracy improvements gradually plateau, while the computational cost keeps increasing, resulting in diminishing returns. We report the accuracy when the number of views is set to 2, as it achieves the most improvement in accuracy with relatively low computational overhead, striking a balance between efficiency and accuracy. For LVB, accuracy initially decreases and then improves. This may be attributed to each view generating divergent answers at first, which causes confusion and consequently requires more views to refine the final answer.

\subsection{Visualization Comparison}

Figure~\ref{fig:score} visualizes the distribution of retrieved frames for the online baseline BOLT and our E-VRAG, based on scores generated by the lightweight VLM. The baseline mainly retrieves frames with scores near the maximum (inside the yellow translucent boxes), while ignoring those with low scores (inside the blue translucent boxes). This results in incomplete frame sampling, with redundancy in high-score regions and missing information in low-score regions. Although higher scores ideally indicate greater relevance, this is not always achieved due to limitations of the scoring model. This highlights the need for a dedicated frame retrieval method rather than simply relying on Top-K selection. In contrast, E-VRAG ensures high-scoring frames are included, while redistributing redundant high-score selections to some lower-score frames. This ensures key information is preserved and provides a more comprehensive video representation. In the case illustrated in Figure~\ref{fig:score}, E-VRAG compensates for the scoring limitations of the lightweight VLM, accurately retrieving frames with low scores and thereby enhancing robustness without sacrificing efficiency.

\subsubsection{Limitation}
Our E-VRAG significantly improves the efficiency of video RAG. However, it still leaves room for further optimization towards real-time video understanding.
Therefore, continuous improvements and research efforts are necessary to further reduce latency, so as to better meet the stringent requirements of real-time video analysis.

\section{Conclusions}
In this paper, we propose E-VRAG, an efficient and accurate RAG-based method for video understanding. The E-VRAG hierarchically decomposes queries to quickly identify potentially relevant frames, and utilizes a lightweight VLM to efficiently score the relationships between queries and frames. As a result, E-VRAG reduces computational costs throughout the entire video RAG process at both data and model levels. To mitigate the potential impact on performance resulting from reduced computational workload, E-VRAG retrieves frames by grouping them according to inter-frame score distribution to emphasize global statistics, rather than relying solely on individual frames. Moreover, E-VRAG further enhances the long video understanding by answering query from multiple complementary views. Extensive experiments across four benchmarks demonstrate that our E-VRAG outperforms baseline methods in both efficiency and accuracy. Furthermore, E-VRAG adopts a plug-and-play design that requires no additional training, and can be seamlessly integrated with future advancements in video VLMs to provide highly efficient solutions.

\bibliography{aaai2026}

\end{document}